\documentclass{article}
\usepackage[margin=1.5in]{geometry}
\usepackage{natbib}
\bibpunct[, ]{(}{)}{,}{a}{}{,}
\def\bibfont{\small}

\usepackage{subcaption}
\usepackage[T1]{fontenc}
\usepackage{lmodern}
\usepackage{listings}
\usepackage{todonotes}

\usepackage{xcolor}

\definecolor{codegreen}{rgb}{0,0.6,0}
\definecolor{codegray}{rgb}{0.5,0.5,0.5}
\definecolor{codepurple}{rgb}{0.58,0,0.82}
\definecolor{backcolor}{rgb}{1.0,1.0,1.0}
\colorlet{punct}{red!60!black}
\definecolor{delim}{RGB}{20,105,176}
\colorlet{numb}{magenta!60!black}

\lstdefinelanguage{json}{
  basicstyle=\normalfont\ttfamily\scriptsize,
  numbers=left,
  numberstyle=\tiny\color{codegray},
  stepnumber=1,
  numbersep=4pt,
  showstringspaces=false,
  breaklines=true,
  backgroundcolor=\color{backcolor},
  literate=
  *{:}{{{\color{punct}{:}}}}{1}
  {,}{{{\color{punct}{,}}}}{1}
  {\{}{{{\color{delim}{\{}}}}{1}
  {\}}{{{\color{delim}{\}}}}}{1}
  {[}{{{\color{delim}{[}}}}{1}
  {]}{{{\color{delim}{]}}}}{1},
}

\usepackage{tikz}
\usetikzlibrary{positioning, decorations.pathmorphing, shapes, arrows}
\usepackage{pgfplots}
\pgfplotsset{width=5cm,compat=1.8}
\usepackage{pgfplotstable}
\pgfmathsetseed{1138}

\usepackage{amsmath}
\usepackage{amsfonts}
\usepackage{url}

\usepackage{comment}
\usepackage{textcomp}
\usepackage{microtype}

\newcommand{\abr}[1]{\textsc{#1}}
\newcommand{\pkg}[1]{\texttt{#1}}
\newcommand{\pyirt}[1]{\texttt{py-irt}}
\newcommand{\irt}[1]{\abr{irt}}
\definecolor{lightblue}{HTML}{3cc7ea}
\definecolor{lightgreen}{HTML}{90EE90}

\newif\ifcomment\commenttrue

\newcommand{\figfile}[1]{p2022_arxiv_py-irt/figures/#1}
\newcommand{\autofig}[1]{p2022_arxiv_py-irt/auto_fig/#1}

\title{\pkg{py-irt}: A Scalable Item Response Theory Library for Python}

\author{
  John P. Lalor\\
  \small IT, Analytics, and Operations \\
  \small University of Notre Dame \\
  \tt{john.lalor@nd.edu}
  \and
  Pedro Rodriguez \\
  \tt{me@pedro.ai}
}
\date{}

\begin{document}
\maketitle
\begin{abstract}

	\pyirt{} is a Python library for fitting Bayesian Item Response Theory (\irt{}) models.
	\pyirt{} estimates latent traits of subjects and items, making it appropriate for use in IRT tasks as well as ideal-point models.
	\pyirt{} is built on top of the Pyro and PyTorch frameworks and uses \abr{gpu}-accelerated training to scale to large data sets.
	Code, documentation, and examples can be found at https://github.com/nd-ball/py-irt.
	\pyirt{} can be installed from the GitHub page or the Python Package Index (PyPI).

\end{abstract}

\section{Introduction}

Item Response Theory (\irt{}) models jointly estimate latent traits of items (e.g., the difficulty of exam questions) and subjects (e.g., the ability of human test-takers).
Originally developed as an alternative to simple summary statistics~\citep{edgeworth1888exams}, \irt{} sees widespread use in education testing\footnote{
    The educational testing field is also known as psychometrics.
}~\citep{lord1968test} for scale construction and evaluation~\citep{carlson2013Item}.
Increasingly, it is also being used to evaluate machine learning models~\citep{lalor2020thesis}.

Research using \irt{} spans fields such as educational testing, machine learning, management, operations, and policital science.
Recent work on \irt{} in machine learning investigates its use for model and data analysis~\citep{martinez-plumed2016Making,martinez-plumed2019Item,vania2021compare}, test set evaluation~\citep{lalor2016Building,rodriguez2021Evaluation}, deep neural network training and behavior~\citep{lalor2019Learning,lalor2020Dynamic}, and chatbot data evaluation~\citep{sedoc2020Item}.

There is also a stream of work in the business community that leverages \irt{} models, including near-miss analyses~\citep{cui2017Polytomous}, scale construction in marketing~\citep{dejong2009Model,dejong2012StateDependence}, and probability forecasting~\citep{satopaa2021Biasa}.
Researchers in operations use \irt{} for machine learning model comparison~\citep{roy2019Performance}.

\irt{} models are known as ideal-point models in political science literature~\citep{gerrish2011predicting,poole2017voting}.
In this context, the models are used to estimate latent party affiliation for politicians based on binary voting records (e.g., voting for or against a proposed bill).
Extensions to ideal point models incorporate the text of the bills and politicians' speeches as well~\citep{nguyen2015tea}.

Existing libraries for fitting \irt{} models are implemented in R and cannot scale to large-scale data.
There is a need for a fast, user-friendly \irt{} library.
We present \pyirt{}, a Python package for fitting Bayesian \irt{} models.
\pyirt{} is built on top of the Pyro probabilistic programming language~\citep{bingham2019pyro}, which itself makes use of PyTorch \abr{gpu}-accelerated tensor math and automatic differentiation~\citep{paszke2019pytorch}.
Models are learned via stochastic variational inference~\citep{kingma2013AutoEncoding} and mini-batch gradient descent, making them scalable to large data sets.

In this work, we describe the development of \pyirt{}, describe the code and its functionality, and demonstrate applications in natural language processing (\abr{nlp}) and computer vision (\abr{cv}).
We demonstrate that \pyirt{} is an easy-to-use and efficient library for developing \irt{} models.

\section{Background}
In this section, we introduce notation for \irt{} models and describe a class of \irt{} models for binary data.
We then demonstrate two methods for estimating \irt{} model parameters: marginal maximum likelihood estimation and variational inference.
Finally, we highlight existing packages for fitting \irt{} models.

\subsection{Notation}
\irt{} is a widely used~\citep{van2016assess} psychometric methodology for learning latent parameters of items and subjects~\citep{baker2001irt,baker2004Item}.
Typically, subjects are human test-takers, and items are assessment questions~\citep[e.g., from a standardized test,][]{dejong2009Model,dejong2012StateDependence}.
For dichotomous data, a subject's responses to a set of items are graded as correct or incorrect.
The binary response matrix of multiple subjects' responses is used to estimate latent parameters of the items and the latent ability of the subjects.\footnote{\irt{} models for polytomous data such as Likert scales also exist.}
\irt{} models learn these latent parameters from the input response pattern matrix $\mathbf{Z}^{I \times J}$, where each row in the matrix represents subject $j's$ responses to the items in the dataset:

\begin{equation}
	Z^j = [\mathbf{I}[\hat{y}^j_0 = y_0], \dots, \mathbb{I}[\hat{y}^j_I = y_I]]
\end{equation}

where $\mathbb{I}$ is the indicator function, and is equal to $1$ when the expression is true and $0$ when the expression is false.
Put another way, $\mathbf{Z}$ is a binarized, graded representation of the subjects' responses to the items.

\subsection{IRT Models}

We next describe several variations of \irt{} models that are implemented in \pkg{py-irt}.

The one-parameter logistic model (1PL), also known as the Rasch model, estimates a latent ability parameter $\theta_j$ for subjects and a latent difficulty parameter $b_i$ for items (Equation~\ref{eqn:1pl}).

\begin{equation}
	\label{eqn:1pl}
	p (y_{ij} = 1 \vert b_i, \theta_j ) = \frac{1}{1+e^{-(\theta_j-b_i)} }
\end{equation}

\begin{figure}
	\centering
	\begin{subfigure}[b]{0.3\textwidth}
		\begin{tikzpicture}
			\begin{axis}[xlabel=$\theta$,ylabel={$p(y_{ij} =1)$}, ymin=0, ymax = 1, xmin=-4, xmax=4]
				\addplot[dashed,black] { ((1) / (1 + exp(-1*(x -2.5))))};
				\addplot[black] { ((1) / (1 + exp(-1*(x +2.5))))};
				\addplot[dashed, black] coordinates{(2.5,0.3) (2.5,0.7)};
				\addplot[solid, black] coordinates{(-2.5,0.3) (-2.5,0.7)};
			\end{axis}
		\end{tikzpicture}
		\caption{\label{fig:1pl}}
	\end{subfigure}
	\begin{subfigure}[b]{0.3\textwidth}
		\begin{tikzpicture}
			\begin{axis}[xlabel=$\theta$,ylabel={$p(y_{ij} =1)$}, ymin=0, ymax = 1, xmin=-4, xmax=4]
				\addplot[dashed,black] { ((1) / (1 + exp(-2*(x -0))))};
				\addplot[black] { ((1) / (1 + exp(-0.5*(x -0))))};

			\end{axis}
		\end{tikzpicture}
		\caption{\label{fig:2pl}}
	\end{subfigure}
	\begin{subfigure}[b]{0.3\textwidth}
		\begin{tikzpicture}
			\begin{axis}[xlabel=$\theta$,ylabel={$p(y_{ij} =1)$}, ymin=0, ymax = 1, xmin=-4, xmax=4]
				\addplot[dashed,black] { 0.2 + ((1-0.2) / (1 + exp(-1*(x -0))))};
				\addplot[black] { 0.05 + ((1-0.05) / (1 + exp(-1*(x +0))))};

			\end{axis}
		\end{tikzpicture}
		\caption{\label{fig:3pl}}
	\end{subfigure}
	\caption{\ref{fig:1pl}: \protect\abr{ICC}s for an easy item (solid) and a hard item (dashed). Vertical lines indicate the point on the ability scale where $p(y_{ij} =1 \vert b_i, \theta_j) = 0.5$. \ref{fig:2pl}: \abr{ICC}s where the discrimination parameter ($a_i$) varies. \ref{fig:3pl}: \abr{ICC}s where the guessing parameter ($c_i$) varies.}
	\label{fig:iccs}
\end{figure}
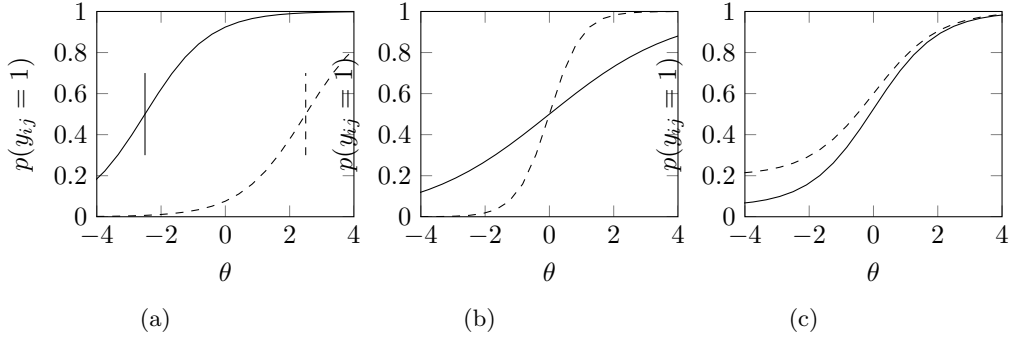

Equation \ref{eqn:1pl} defines an Item Characteristic Curve (\abr{icc}s) for a particular item (Figure~\ref{fig:iccs}).
For a particular subject, the probability of correctly answering a given item is a function of the subject's latent ability and the item's latent characteristics.
For individuals of higher ability, the probability of correctly answering an item increases---all else equal.

More complex \irt{} models estimate additional latent parameters for the items.
The two-parameter logistic (2PL) model adds a discrimination parameter $a_i$, so that different items have steeper or shallower slopes at the steepest point (Equation \ref{eqn:2pl}).
The three-parameter logistic (3PL) model adds a guessing parameter ($c_i$) to model the fact that, even at low levels of subject ability, there is a non-zero probability that a subject will guess a question's answer correctly (Equation~\ref{eqn:3pl}).
The feasibility model of~\citet{rodriguez2021Evaluation} sets an upper bound on the \abr{icc} for ``unanswerable'' questions (Equation~\ref{eqn:4pl}).
These models can also be multidimensional in the latent space~\citep{reckase2009mirt}.

\begin{align} 
\label{eqn:2pl}
p (y_{ij} = 1 \vert a_i, b_i, \theta_j )&= \frac{1}{1+e^{-a_i(\theta_j-b_i)} }\\
\label{eqn:3pl}
p (y_{ij} = 1 \vert a_i, b_i, c_i, \theta_j )&= c_i + \frac{1 - c_i}{1+e^{-a_i(\theta_j-b_i)} }\\
\label{eqn:4pl}
p (y_{ij} = 1 \vert a_i, b_i, \lambda_i, \theta_j )&= \frac{\lambda_i}{1+e^{-a_i(\theta_j-b_i)} }
\end{align}

\subsection{Parameter Estimation}

To derive model parameters, we use probabilistic inference~\citep{pearl-88} to find the latent subject and item parameters that best explain the data---in this case the response pattern matrix $\mathbf{Z}$.
The likelihood of a particular response matrix given latent subject and item parameters is:\footnote{In this section, we describe parameter estimation for 1PL models, but the process is the same for more advanced models.} 

\begin{equation}
	p(\mathbf{Z} \vert \Theta, B) = \prod_{j=1}^{J} \prod_{i=1}^{I} p(\mathbf{Z}_{ij} = y_{ij} \vert \theta_j, b_i).
\end{equation}

For computational efficiency, this is often represented in log space:

\begin{equation}
	\log p(\mathbf{Z} \vert \Theta, B) = \sum_{j=1}^{J} \sum_{i=1}^{I} \log p(\mathbf{Z}_{ij} = y_{ij} \vert \theta_j, b_i).
\end{equation}

Traditionally, this is fit using marginal maximum likelihood estimation.
\cite{bock1981Marginal} proposed an expectation-maximization algorithm, where ability is integrated out, and at each time step $t$ the item parameters are estimated using maximum a posteriori (MAP) estimation:

\begin{align}
	\text{E step: } \quad     &p(z_{ij} \vert b_i^t)   = \int_{\theta_i} p(z_{ij} \vert \theta_j, b_i^t)p(\theta_j)d\theta_j\\
	\text{M step: } \quad     &b_i^{t+1} = \arg\max_{b_i} \sum_{i}^{I}\log p(z_{ij} \vert b_i^t)
\end{align}

Once item parameter estimation is complete, estimating subject ability is also done via MAP estimation.

Marginal maximum likelihood estimation does not scale well to large data sets or a large number of subjects.
As an alternative, some have proposed Bayesian methods to account both for scaling and also to allow for uncertainty estimation~\citep{natesan2016Bayesian}.
The Bayesian approach attempts to estimate the latent parameters given the response pattern data:

\begin{equation}
	p(\Theta, B \vert \mathbf{Z})
\end{equation}

Because calculating this posterior is intractable, approximate inference techniques such as variational inference are used instead~\citep{jordan1998Introduction}.

Recent work has investigated using variational inference (\abr{vi})~\citep{natesan2016Bayesian}.
\abr{vi} estimates a distribution over the latent variables that approximates the true posterior and considers mean and variance estimates for calculating uncertainty.
Means for the variational parameters are drawn from Normal distributions, and scale parameters are drawn from Gamma distributions~\citep{lalor2019Learning}.

\begin{equation}
	q_{\phi}(\theta, b, \mu, \tau) = q(\mu)q(\tau) \prod_{i,j} q(\theta_j)q(b_i)
\end{equation}

Now the optimization task is to identify the variational parameters that minimize the \abr{kl} Divergence ($D_{\text{KL}}$) with the true (unknown) posterior:

\begin{equation}
		q_{\phi^*}(\theta, b, \mu, \tau) =\arg \min_{q_\phi} D_{\text{KL}}(q_{\phi}(\theta, b, \mu, \tau)||p(\theta, b \vert \mathbf{Z}))
\end{equation}

\subsection{Related Libraries}

Most \irt{}-related software packages are implemented in R and target human-sized data, where the number of subjects and items is relatively small.\footnote{Collecting large-scale human data is difficult since human subject recruitment is expensive, and humans can only answer a relatively small number of items for a test.}
However, when applying \irt{} models to machine learning-scale problems \citep[e.g., an ensemble of neural network models trained on the 50,000 \abr{cifar} training examples,][]{krizhevsky2009learning}, many existing \irt{} programs are not able to scale despite work showing that \abr{vi} makes this possible~\citep{wu2020virt}.
For researchers working in Python with machine learning libraries such as Tensorflow \citep{abadi2016tensorflow} or PyTorch \citep{paszke2019pytorch,zhang2021LearningBased}, moving from Python to R to incorporate \irt{} analysis is not ideal.
To the best of our knowledge, there is no existing resource for the research community to fit large \irt{} models in Python, particularly Bayesian \irt{} models that allow for estimating uncertainty in parameter estimation.

Existing packages for fitting \irt{} models include \pkg{mirt}~\citep{chalmers2015Mirt}, \pkg{ltm}~\citep{rizopoulos2006Ltm}, and \pkg{equateIRT}~\citep{battauz2015equateirt} in R.
\irt{} models can also be fit in Python using Stan~\citep{carpenter2017stan}, but this requires users to write down their model using the Stan configuration and due to its use of Markov Chain Monte Carlo methods~\citep{kim2007irtmc} takes substantially longer to run.
There is a need for a user-friendly \irt{} library in Python, particularly for users in the wider research community who may want to easily plug difficulty and ability estimation into a larger (typically Pythonic) data science pipeline. 

\pkg{py-irt} has several benefits: 1) as a high-level Python package, users can easily fit \irt{} models for their data sets, 2) \pkg{py-irt} can be used as a command-line interface (CLI) or Python module for flexibility, 3) \pkg{py-irt}'s Bayesian implementation allows for uncertainty in output via latent trait means and variances, and 4) by leveraging \pkg{pyro}'s PyTorch backend, large-scale models can be fit efficiently via \abr{gpu} acceleration.
The scalability and ease of use make \pkg{py-irt} beneficial for researchers looking to implement larger models with larger data sets of items and subjects.

\section{Implementation}

We next describe how \pkg{py-irt} is implemented and demonstrate how researchers can fit \irt{} models and define new \irt{} models in \pkg{py-irt}.
\pkg{py-irt} is implemented in Python and built using Pyro, a probabilistic programming language~\citep{bingham2019pyro}.
Pyro allows for the flexible creation of Bayesian models.
To achieve scalability, it uses a PyTorch~\citep{paszke2019pytorch} backend.
Pyro implements several inference algorithms, such as \abr{nuts}~\citep{hoffman2014no} and stochastic variational inference~\citep{kingma2013AutoEncoding}.

\subsection{Fitting Models}

For practitioners, \pkg{py-irt} can be used as either a command-line interface (CLI) or a Python module.
To run \pkg{py-irt}, users first format their response pattern dataset into the \pyirt{} jsonlines format (Figure~\ref{fig:jsonlines}).
Each row in the data represents a different subject.
Each subject has a unique \abr{id}, and a dictionary of responses ($1$ or $0$) for uniquely identified items.
This maintains readability of the dataset while still allowing for sparse response patterns.

\begin{figure}[h!]
  \centering
  \begin{subfigure}[b]{0.45\textwidth}
  \lstinputlisting[language=json]{\figfile{minitest.jsonlines}}
  \caption{\label{fig:jsonlines}}
  \end{subfigure}
  \begin{subfigure}[b]{0.45\textwidth}
  \lstinputlisting[language=json]{\figfile{best_parameters.json}}
  \caption{\label{fig:output}}
  \end{subfigure}
  \caption{An example of the \pkg{py-irt} input data format (\ref{fig:jsonlines}) and the \pkg{py-irt} output for a 1PL model (\ref{fig:output}).}
\end{figure}

To fit a 1PL model from the command line, users execute a single command (Figure \ref{fig:cliTrain}).\footnote{\irt{} models can also be fit within Python code. To do this, the user loads the input data, followed by initializing and running the desired model. See for example: \url{https://github.com/nd-ball/py-irt/blob/master/examples/example_with_rps.py}.}
The output is a \abr{json} dictionary containing estimated item and subject parameters.
Since subject and item identifiers are not tied to their positions in the model, the output file also links the position of parameters and \abr{id}s~(Figure~\ref{fig:output}).

\begin{figure}[h!]
\begin{lstlisting}[language=bash, basicstyle=\small]
	py-irt train 1pl data/data.jsonlines output/1pl/
\end{lstlisting}
\caption{Command line function for training a \pkg{py-irt} model.}
\label{fig:cliTrain}
\end{figure}

\subsection{Defining Models}

Researchers can define their own \irt{} models using \pkg{py-irt}.
In the Pyro tradition, \pyirt{} implements models using the model-guide paradigm.
The probabilistic model is specified, and variational guide distributions are initialized~(Figure~\ref{fig:pyro}).

\begin{figure}[h!]
\begin{subfigure}[b]{0.9\textwidth}
  \begin{lstlisting}[language=Python, basicstyle=\small]
	@abstract_model.IrtModel.register("new1PL")
	class NewOneParamLog(abstract_model.IrtModel):
\end{lstlisting}
\caption{\label{fig:newModel}}
\end{subfigure}
\begin{subfigure}[b]{0.9\textwidth}
\begin{lstlisting}[language=bash, basicstyle=\small]
	py-irt train new1PL data/data.jsonlines output/new1PL/
\end{lstlisting}
\caption{\label{fig:cliNew1PL}}
\end{subfigure}
\caption{\ref{fig:newModel}. Python example of registering a new \abr{irt} model with \pkg{py-irt}. \ref{fig:cliNew1PL}. Command line code for training the new1PL model.}
\end{figure}

\begin{figure}[h!]
  \centering
  \begin{subfigure}[b]{0.9\textwidth}
    \centering
    \lstinputlisting[language=Python]{\figfile{onePL_model.py}}
    
    \caption{\label{fig:model}}
  \end{subfigure}
  \begin{subfigure}[b]{0.9\textwidth}
    \centering
    \lstinputlisting[language=Python]{\figfile{onePL_guide.py}}
    
    \caption{\label{fig:guide}}
  \end{subfigure}
  \caption{\ref{fig:model}: In the Pyro model, developers initialize the latent parameters and define the sampling statements. \ref{fig:guide}: The guide distributions estimate the mean and variance parameters approximating the model distribution.}
  \label{fig:pyro}
\end{figure}

Once the developer creates and registers the model, it is available from the \abr{cli}.\footnote{
	This registering pattern is inspired by Allen\abr{nlp}~\citep{gardner2017allennlp}.
}
For example, a user can register the ``NewOneParamLog'' \irt{} model as an instance of IrtModel (Figure \ref{fig:newModel}).
To train this model using the \abr{cli} on our example data set involves changing the model specification (Figure \ref{fig:cliNew1PL}).
With \pkg{py-irt}, researchers can easily incorporate implemented \irt{} models into their data analysis, and can define and fit new \irt{} models within the library.

\section{Experiments and Examples}

\begin{figure}
	\centering
	\includegraphics[width=\linewidth]{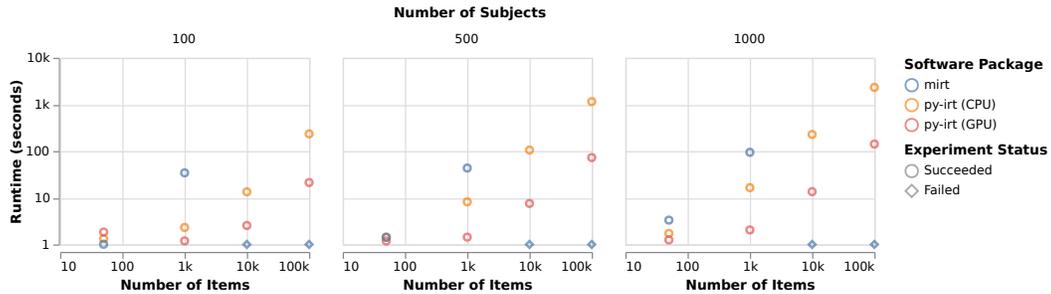}
	\caption{Runtime comparison to fit a 1PL model with varying levels of items and subjects.
	Beyond 1000 items, \pkg{mirt} does not reliably complete training.}
	\label{fig:runtime}
\end{figure}

To demonstrate the speed of \pkg{py-irt} we compare fitting 1PL models using \pkg{py-irt} with \pkg{mirt}, a popular R package for \irt{}.
We randomly initialize response pattern matrices of different sizes and compare the runtime to fit \irt{} models in both packages.
We vary the number of items and the number of subjects.
We do this to simulate \irt{} use-case scenarios, such as fitting a model for a single human user-study with a relatively small number of items and subjects, to a machine learning evaluation where the number of subjects and items can both be very large.
From this, we plot how runtime changes as the number of items and/or subjects increases (Figure~\ref{fig:runtime}).
As the data matrix size grows, \pkg{mirt} takes much longer than \pyirt{} to fit the model.
Beyond 1000 items, \pkg{mirt} is not able to fit an \irt{} model.
\pkg{py-irt} can fit much larger models.
The \abr{gpu} acceleration made available by PyTorch further accelerates training to under a minute even for large data sets (i.e., more than 100,000 items and 100 subjects).

In the next experiment, we demonstrate that \pyirt{}-derived parameters are also reasonable across computer vision and natural language processing data sets.

1PL models are fit for the \abr{mnist}~\citep{lecun1998MNIST} and \abr{cifar}~\citep{krizhevsky2009learning} computer vision data sets and the Stanford Sentiment Treebank natural language processing data set~\citep{socher2013recursive}.
The \irt{} models are fit using response pattern data from an ensemble of machine learning models~\citep{lalor2019Learning}.
These data sets are widely used in machine learning research and see use in operations research as well~\citep[e.g., ][]{chen2021ClusterAware}.
 
\begin{figure}[th!]
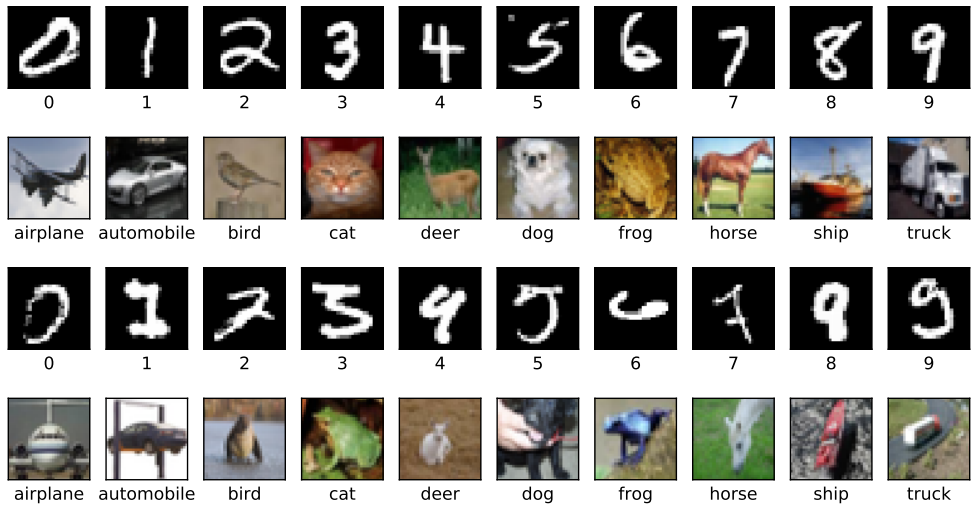

	\centering
	\begin{subfigure}[b]{\textwidth}
		\centering
		\includegraphics[width=0.95\linewidth]{\figfile{easiest_MNIST_test}}
		\caption{\label{sfig1a}}
		\vspace{-2em}
	\end{subfigure}
	\begin{subfigure}[b]{\textwidth}
		\centering
		\includegraphics[width=0.95\linewidth]{\figfile{easiest_CIFAR_test}}
		\caption{\label{sfig1c}} 
		\vspace{-2em}
	\end{subfigure}
	\begin{subfigure}[b]{\textwidth}
		\centering
		\includegraphics[width=0.95\linewidth]{\figfile{hardest_MNIST_test}}
		\caption{\label{sfig1b}} 
		\vspace{-2em}
	\end{subfigure}
	\begin{subfigure}[b]{\textwidth}
		\centering
		\includegraphics[width=0.95\linewidth]{\figfile{hardest_CIFAR_test}}
		\caption{\label{sfig1d}} 
		\vspace{-2em}
	\end{subfigure}
	\caption{(Best viewed in color) The easiest (first and second rows) and hardest (third and fourth rows) items in the MNIST and CIFAR test sets. Images from \citep{lalor2019Learning}.}
	\label{fig:vizexamples}
\end{figure}

In all cases, the items in the data sets identified as easy and hard seem appropriate upon inspection (Figure~\ref{fig:vizexamples} and Table~\ref{tab:sstb}).
The easy and difficult items in both cases are interpretable and intuitive and show how difficulty can manifest as a rare example (e.g., the blue frog), an incorrect label (e.g., the frog labeled cat), or as a case near a decision boundary (e.g., the slightly negative review). 
Having detailed information on item difficulty can inform data set curation and make model predictions more interpretable~\citep{baldock2021deep}; knowing latent item parameters also means that machine learning model ability can be estimated for more detailed analyses of model performance.

\captionsetup[subfigure]{labelformat=empty}

\begin{table}[h]
  \small
  \centering 
  \begin{tabular}{p{9cm}ll}
    \hline \bf Phrase & \bf Label  & \bf Difficulty\\ \hline
    The stupidest, most insulting movie of 2002's first quarter. & Negative & -2.46 \\
    An endlessly fascinating, landmark movie that is as bold as anything the cinema has seen in years. & Positive & -2.27\\
    Still, it gets the job done - a sleepy afternoon rental. & Negative & 1.78 \\
    Perhaps no picture ever made has more literally showed that the road to hell is paved with good intentions. & Positive  & 2.05 \\ \hline
  \end{tabular}
  \caption{Easiest and hardest items from a sentiment classification task as estimated by a 1PL model.}
  \label{tab:sstb}
\end{table}

\section{Conclusion} 

In this work, we introduced \pyirt{}, a Python package for Bayesian Item Response Theory modeling.
\pyirt{} can be used to easily and efficiently fit \irt{} models, ideal point models, and other latent trait models.
The scalability of \pyirt{}---owing to its roots in \abr{gpu}-accelerated PyTorch and Pyro---allows for investigation of larger datasets like those used in machine learning evaluations.
The package is available online and set up for code contributions as well.
Future development for the package includes implementing amortized latent trait models and adding graded response models that can be used for Likert-scale type responses.

The code for \pyirt{} is freely available at http://www.github.com/nd-ball/py-irt under the \abr{mit} license.
The Github page includes examples, an issue-tracker, and instructions for community members to contribute.
Continuous integration testing ensures that new code is automatically tested before being merged into the codebase. 
Documentation is available at https://readthedocs.org/projects/py-irt/.
The package is also available for download via the Python Package Index (PyPI): https://pypi.org/project/py-irt/.

\bibliographystyle{style/informs2014}
\bibliography{bib/journal-full,bib/2021_IJoC_Pyirt}

\begin{thebibliography}{46}
\providecommand{\natexlab}[1]{#1}
\providecommand{\url}[1]{\texttt{#1}}
\providecommand{\urlprefix}{URL }

\bibitem[{Abadi et~al.(2016)Abadi, Agarwal, Barham, Brevdo, Chen, Citro,
  Corrado, Davis, Dean, Devin et~al.}]{abadi2016tensorflow}
Abadi M, Agarwal A, Barham P, Brevdo E, Chen Z, Citro C, Corrado GS, Davis A,
  Dean J, Devin M, et~al. (2016) Tensorflow: {{Large-scale}} machine learning
  on heterogeneous distributed systems. \emph{arXiv preprint arXiv:1603.04467}
  .

\bibitem[{Baker(2001)}]{baker2001irt}
Baker FB (2001) \emph{The Basics of Item Response Theory} (ERIC), ISBN
  ISBN-1-886047-03-0.

\bibitem[{Baker \protect\BIBand{} Kim(2004)}]{baker2004Item}
Baker FB, Kim SH (2004) \emph{Item {{Response Theory}}: {{Parameter Estimation
  Techniques}}, {{Second Edition}}} ({CRC Press}), ISBN 978-0-8247-5825-7.

\bibitem[{Baldock et~al.(2021)Baldock, Maennel, \protect\BIBand{}
  Neyshabur}]{baldock2021deep}
Baldock R, Maennel H, Neyshabur B (2021) Deep learning through the lens of
  example difficulty. \emph{Advances in Neural Information Processing Systems}
  34.

\bibitem[{Battauz(2015)}]{battauz2015equateirt}
Battauz M (2015) {{equateIRT}}: {{An}} r package for {{IRT}} test equating.
  \emph{Journal of Statistical Software} 68(1):1--22.

\bibitem[{Bingham et~al.(2019)Bingham, Chen, Jankowiak, Obermeyer, Pradhan,
  Karaletsos, Singh, Szerlip, Horsfall, \protect\BIBand{}
  Goodman}]{bingham2019pyro}
Bingham E, Chen JP, Jankowiak M, Obermeyer F, Pradhan N, Karaletsos T, Singh R,
  Szerlip P, Horsfall P, Goodman ND (2019) Pyro: {{Deep}} universal
  probabilistic programming. \emph{The Journal of Machine Learning Research}
  20(1):973--978.

\bibitem[{Bock \protect\BIBand{} Aitkin(1981)}]{bock1981Marginal}
Bock RD, Aitkin M (1981) Marginal maximum likelihood estimation of item
  parameters: {{Application}} of an {{EM}} algorithm. \emph{Psychometrika}
  46(4):443--459.

\bibitem[{Carlson \protect\BIBand{} von Davier(2013)}]{carlson2013Item}
Carlson JE, von Davier M (2013) Item {{Response Theory}}. \emph{ETS Research
  Report Series} 2013(2):i--69, ISSN 2330-8516,
  \urlprefix\url{http://dx.doi.org/10.1002/j.2333-8504.2013.tb02335.x}.

\bibitem[{Carpenter et~al.(2017)Carpenter, Gelman, Hoffman, Lee, Goodrich,
  Betancourt, Brubaker, Guo, Li, \protect\BIBand{} Riddell}]{carpenter2017stan}
Carpenter B, Gelman A, Hoffman MD, Lee D, Goodrich B, Betancourt M, Brubaker M,
  Guo J, Li P, Riddell A (2017) Stan: {{A}} probabilistic programming language.
  \emph{Journal of statistical software} 76(1):1--32.

\bibitem[{Chalmers et~al.(2015)Chalmers, Pritikin, Robitzsch, \protect\BIBand{}
  Zoltak}]{chalmers2015Mirt}
Chalmers P, Pritikin J, Robitzsch A, Zoltak M (2015) \emph{Mirt:
  {{Multidimensional Item Response Theory}}}.

\bibitem[{Chen \protect\BIBand{} Xie(2021)}]{chen2021ClusterAware}
Chen S, Xie W (2021) On {{Cluster-Aware Supervised Learning}}: {{Frameworks}},
  {{Convergent Algorithms}}, and {{Applications}}. \emph{INFORMS Journal on
  Computing} ISSN 1091-9856,
  \urlprefix\url{http://dx.doi.org/10.1287/ijoc.2020.1053}.

\bibitem[{Cui et~al.(2017)Cui, Rosoff, \protect\BIBand{}
  John}]{cui2017Polytomous}
Cui J, Rosoff H, John RS (2017) A {{Polytomous Item Response Theory Model}} for
  {{Measuring Near-Miss Appraisal}} as a {{Psychological Trait}}.
  \emph{Decision Analysis} 14(2):75--86, ISSN 1545-8490, 1545-8504,
  \urlprefix\url{http://dx.doi.org/10.1287/deca.2017.0345}.

\bibitem[{{de Jong} et~al.(2012){de Jong}, Lehmann, \protect\BIBand{}
  Netzer}]{dejong2012StateDependence}
{de Jong} MG, Lehmann DR, Netzer O (2012) State-{{Dependence Effects}} in
  {{Surveys}}. \emph{Marketing Science} 31(5):838--854, ISSN 0732-2399,
  1526-548X, \urlprefix\url{http://dx.doi.org/10.1287/mksc.1120.0722}.

\bibitem[{{de Jong} et~al.(2009){de Jong}, Steenkamp, \protect\BIBand{}
  Veldkamp}]{dejong2009Model}
{de Jong} MG, Steenkamp JBEM, Veldkamp BP (2009) A {{Model}} for the
  {{Construction}} of {{Country-Specific Yet Internationally Comparable
  Short-Form Marketing Scales}}. \emph{Marketing Science} 28(4):674--689, ISSN
  0732-2399, 1526-548X,
  \urlprefix\url{http://dx.doi.org/10.1287/mksc.1080.0439}.

\bibitem[{Edgeworth(1888)}]{edgeworth1888exams}
Edgeworth FY (1888) The statistics of examinations. \emph{Journal of the Royal
  Statistical Society} 51(3):599--635, ISSN 0952-8385,
  \urlprefix\url{http://www.jstor.org/stable/2339898}.

\bibitem[{Gardner et~al.(2018)Gardner, Grus, Neumann, Tafjord, Dasigi, Liu,
  Peters, Schmitz, \protect\BIBand{} Zettlemoyer}]{gardner2017allennlp}
Gardner M, Grus J, Neumann M, Tafjord O, Dasigi P, Liu NF, Peters M, Schmitz M,
  Zettlemoyer LS (2018) Allennlp: A deep semantic natural language processing
  platform.

\bibitem[{Gerrish \protect\BIBand{} Blei(2011)}]{gerrish2011predicting}
Gerrish SM, Blei DM (2011) Predicting legislative roll calls from text.
  \emph{Proceedings of the 28th International Conference on Machine Learning,
  {{ICML}} 2011}.

\bibitem[{Hoffman et~al.(2014)Hoffman, Gelman et~al.}]{hoffman2014no}
Hoffman MD, Gelman A, et~al. (2014) The no-u-turn sampler: Adaptively setting
  path lengths in hamiltonian monte carlo. \emph{Journal of Machine Learning
  Research} 15(1):1593--1623.

\bibitem[{Jordan et~al.(1998)Jordan, Ghahramani, Jaakkola, \protect\BIBand{}
  Saul}]{jordan1998Introduction}
Jordan MI, Ghahramani Z, Jaakkola TS, Saul LK (1998) An {{Introduction}} to
  {{Variational Methods}} for {{Graphical Models}}. Jordan MI, ed.,
  \emph{Learning in {{Graphical Models}}}, 105--161, {{NATO ASI Series}}
  ({Dordrecht}: {Springer Netherlands}), ISBN 978-94-011-5014-9,
  \urlprefix\url{http://dx.doi.org/10.1007/978-94-011-5014-9_5}.

\bibitem[{Kim \protect\BIBand{} Bolt(2007)}]{kim2007irtmc}
Kim J, Bolt D (2007) Estimating item response theory models using markov chain
  monte carlo methods. \emph{Educational Measurement: Issues and Practice}
  26:38 -- 51,
  \urlprefix\url{http://dx.doi.org/10.1111/j.1745-3992.2007.00107.x}.

\bibitem[{Kingma \protect\BIBand{} Welling(2013)}]{kingma2013AutoEncoding}
Kingma DP, Welling M (2013) Auto-{{Encoding Variational Bayes}}.
  \emph{arXiv:1312.6114 [cs, stat]} .

\bibitem[{Krizhevsky et~al.(2009)}]{krizhevsky2009learning}
Krizhevsky A, et~al. (2009) Learning multiple layers of features from tiny
  images .

\bibitem[{Lalor(2020)}]{lalor2020thesis}
Lalor JP (2020) \emph{Learning Latent Characteristics of Data and Models using
  Item Response Theory}. Ph.D. thesis, University of Massachusetts Amherst,
  \urlprefix\url{http://dx.doi.org/10.7275/reha-je40}.

\bibitem[{Lalor et~al.(2016)Lalor, Wu, \protect\BIBand{}
  Yu}]{lalor2016Building}
Lalor JP, Wu H, Yu H (2016) Building an {{Evaluation Scale}} using {{Item
  Response Theory}}. \emph{Proceedings of the {{Conference}} on {{Empirical
  Methods}} in {{Natural Language Processing}}. {{Conference}} on {{Empirical
  Methods}} in {{Natural Language Processing}}}, volume 2016, 648--657.

\bibitem[{Lalor et~al.(2019)Lalor, Wu, \protect\BIBand{}
  Yu}]{lalor2019Learning}
Lalor JP, Wu H, Yu H (2019) Learning {{Latent Parameters}} without {{Human
  Response Patterns}}: {{Item Response Theory}} with {{Artificial Crowds}}.
  \emph{Proceedings of the {{Conference}} on {{Empirical Methods}} in {{Natural
  Language Processing}}. {{Conference}} on {{Empirical Methods}} in {{Natural
  Language Processing}}}, volume 2019 ({Association for Computational
  Linguistics}).

\bibitem[{Lalor \protect\BIBand{} Yu(2020)}]{lalor2020Dynamic}
Lalor JP, Yu H (2020) Dynamic {{Data Selection}} for {{Curriculum Learning}}
  via {{Ability Estimation}}. \emph{Findings of the {{Association}} for
  {{Computational Linguistics}}: {{EMNLP}} 2020}, 545--555 ({Online}:
  {Association for Computational Linguistics}),
  \urlprefix\url{http://dx.doi.org/10.18653/v1/2020.findings-emnlp.48}.

\bibitem[{LeCun et~al.(1998)LeCun, Cortes, \protect\BIBand{}
  Burges}]{lecun1998MNIST}
LeCun Y, Cortes C, Burges CJ (1998) {{MNIST}} handwritten digit database.
  http://yann.lecun.com/exdb/mnist/.

\bibitem[{Lord et~al.(1968)Lord, Novick, \protect\BIBand{}
  Birnbaum}]{lord1968test}
Lord FM, Novick MR, Birnbaum A (1968) Statistical theories of mental test
  scores \urlprefix\url{https://psycnet.apa.org/fulltext/1968-35040-000.pdf}.

\bibitem[{{Mart{\'i}nez-Plumed} et~al.(2019){Mart{\'i}nez-Plumed},
  Prud{\^e}ncio, {Mart{\'i}nez-Us{\'o}}, \protect\BIBand{}
  {Hern{\'a}ndez-Orallo}}]{martinez-plumed2019Item}
{Mart{\'i}nez-Plumed} F, Prud{\^e}ncio RBC, {Mart{\'i}nez-Us{\'o}} A,
  {Hern{\'a}ndez-Orallo} J (2019) Item response theory in {{AI}}: {{Analysing}}
  machine learning classifiers at the instance level. \emph{Artificial
  Intelligence} 271:18--42, ISSN 0004-3702,
  \urlprefix\url{http://dx.doi.org/10.1016/j.artint.2018.09.004}.

\bibitem[{{Mart{\'i}nez-Plumed} et~al.(2016){Mart{\'i}nez-Plumed},
  Prud{\^e}ncio, Us{\'o}, \protect\BIBand{}
  {Hern{\'a}ndez-Orallo}}]{martinez-plumed2016Making}
{Mart{\'i}nez-Plumed} F, Prud{\^e}ncio RBC, Us{\'o} AM, {Hern{\'a}ndez-Orallo}
  J (2016) Making {{Sense}} of {{Item Response Theory}} in {{Machine
  Learning}}. \emph{{{ECAI}}}, volume 285 of \emph{Frontiers in {{Artificial
  Intelligence}} and {{Applications}}}, 1140--1148 ({IOS Press}),
  \urlprefix\url{http://dx.doi.org/10.3233/978-1-61499-672-9-1140}.

\bibitem[{Natesan et~al.(2016)Natesan, Nandakumar, Minka, \protect\BIBand{}
  Rubright}]{natesan2016Bayesian}
Natesan P, Nandakumar R, Minka T, Rubright JD (2016) Bayesian {{Prior Choice}}
  in {{IRT Estimation Using MCMC}} and {{Variational Bayes}}. \emph{Frontiers
  in Psychology} 7, ISSN 1664-1078,
  \urlprefix\url{http://dx.doi.org/10.3389/fpsyg.2016.01422}.

\bibitem[{Nguyen et~al.(2015)Nguyen, {Boyd-Graber}, Resnik, \protect\BIBand{}
  Miler}]{nguyen2015tea}
Nguyen VA, {Boyd-Graber} J, Resnik P, Miler K (2015) Tea party in the house:
  {{A}} hierarchical ideal point topic model and its application to republican
  legislators in the 112th congress. \emph{Proceedings of the 53rd Annual
  Meeting of the Association for Computational Linguistics and the 7th
  International Joint Conference on Natural Language Processing (Volume 1:
  {{Long}} Papers)}, 1438--1448.

\bibitem[{Paszke et~al.(2019)Paszke, Gross, Massa, Lerer, Bradbury, Chanan,
  Killeen, Lin, Gimelshein, Antiga et~al.}]{paszke2019pytorch}
Paszke A, Gross S, Massa F, Lerer A, Bradbury J, Chanan G, Killeen T, Lin Z,
  Gimelshein N, Antiga L, et~al. (2019) Pytorch: {{An}} imperative style,
  high-performance deep learning library. \emph{Advances in neural information
  processing systems} 32:8026--8037.

\bibitem[{Pearl(1988)}]{pearl-88}
Pearl J (1988) \emph{Probabilistic Reasoning in Intelligent Systems: Networks
  of Plausible Inference} (San Francisco, CA, USA: Morgan Kaufmann Publishers
  Inc.), ISBN 1558604790.

\bibitem[{Poole \protect\BIBand{} Rosenthal(2017)}]{poole2017voting}
Poole KT, Rosenthal H (2017) \emph{Ideology \& congress: A political economic
  history of roll call voting} (London, England: Routledge), 2 edition, ISBN
  9780203789223, \urlprefix\url{http://dx.doi.org/10.4324/9780203789223}.

\bibitem[{Reckase(2009)}]{reckase2009mirt}
Reckase MD (2009) Multidimensional item response theory models. {Reckase}, ed.,
  \emph{Multidimensional Item Response Theory}, 79--112 (New York, NY: Springer
  New York), ISBN 9780387899763,
  \urlprefix\url{http://dx.doi.org/10.1007/978-0-387-89976-3\_4}.

\bibitem[{Rizopoulos(2006)}]{rizopoulos2006Ltm}
Rizopoulos D (2006) Ltm: {{An R Package}} for {{Latent Variable Modeling}} and
  {{Item Response Analysis}}. \emph{Journal of Statistical Software, Articles}
  17(5):1--25, ISSN 1548-7660,
  \urlprefix\url{http://dx.doi.org/10.18637/jss.v017.i05}.

\bibitem[{Rodriguez et~al.(2021)Rodriguez, Barrow, Hoyle, Lalor, Jia,
  \protect\BIBand{} {Boyd-Graber}}]{rodriguez2021Evaluation}
Rodriguez P, Barrow J, Hoyle AM, Lalor JP, Jia R, {Boyd-Graber} J (2021)
  Evaluation {{Examples}} are not {{Equally Informative}}: {{How}} should that
  change {{NLP Leaderboards}}? \emph{Proceedings of the 59th {{Annual Meeting}}
  of the {{Association}} for {{Computational Linguistics}} and the 11th
  {{International Joint Conference}} on {{Natural Language Processing}}
  ({{Volume}} 1: {{Long Papers}})}, 4486--4503 ({Online}: {Association for
  Computational Linguistics}),
  \urlprefix\url{http://dx.doi.org/10.18653/v1/2021.acl-long.346}.

\bibitem[{Roy et~al.(2019)Roy, Qureshi, Pande, Nair, Gairola, Jain, Singh,
  Sharma, Jagadale, Lin, Sharma, Gotety, Zhang, Tang, Mehta, Sindhanuru,
  Okafor, Das, Gopal, Rudraraju, \protect\BIBand{}
  Kakarlapudi}]{roy2019Performance}
Roy A, Qureshi S, Pande K, Nair D, Gairola K, Jain P, Singh S, Sharma K,
  Jagadale A, Lin YY, Sharma S, Gotety R, Zhang Y, Tang J, Mehta T, Sindhanuru
  H, Okafor N, Das S, Gopal CN, Rudraraju SB, Kakarlapudi AV (2019) Performance
  {{Comparison}} of {{Machine Learning Platforms}}. \emph{INFORMS Journal on
  Computing} 31(2):207--225, ISSN 1091-9856,
  \urlprefix\url{http://dx.doi.org/10.1287/ijoc.2018.0825}.

\bibitem[{Satop{\"a}{\"a} et~al.(2021)Satop{\"a}{\"a}, Salikhov, Tetlock,
  \protect\BIBand{} Mellers}]{satopaa2021Biasa}
Satop{\"a}{\"a} VA, Salikhov M, Tetlock PE, Mellers B (2021) Bias,
  {{Information}}, {{Noise}}: {{The BIN Model}} of {{Forecasting}}.
  \emph{Management Science} mnsc.2020.3882, ISSN 0025-1909, 1526-5501,
  \urlprefix\url{http://dx.doi.org/10.1287/mnsc.2020.3882}.

\bibitem[{Sedoc \protect\BIBand{} Ungar(2020)}]{sedoc2020Item}
Sedoc J, Ungar L (2020) Item {{Response Theory}} for {{Efficient Human
  Evaluation}} of {{Chatbots}}. \emph{Proceedings of the {{First Workshop}} on
  {{Evaluation}} and {{Comparison}} of {{NLP Systems}}}, 21--33 ({Online}:
  {Association for Computational Linguistics}),
  \urlprefix\url{http://dx.doi.org/10.18653/v1/2020.eval4nlp-1.3}.

\bibitem[{Socher et~al.(2013)Socher, Perelygin, Wu, Chuang, Manning, Ng,
  \protect\BIBand{} Potts}]{socher2013recursive}
Socher R, Perelygin A, Wu J, Chuang J, Manning DC, Ng A, Potts C (2013)
  Recursive {{Deep Models}} for {{Semantic Compositionality Over}} a
  {{Sentiment Treebank}}. \emph{Proceedings of the 2013 {{Conference}} on
  {{Empirical Methods}} in {{Natural Language Processing}}}, 1631--1642
  ({Seattle, Washington, USA}: {Association for Computational Linguistics}).

\bibitem[{van Rijn et~al.(2016)van Rijn, Sinharay, Haberman, \protect\BIBand{}
  Johnson}]{van2016assess}
van Rijn PW, Sinharay S, Haberman SJ, Johnson MS (2016) Assessment of fit of
  item response theory models used in large-scale educational survey
  assessments. \emph{Large-scale Assessments in Education} 4(1):10, ISSN
  2196-0739, \urlprefix\url{http://dx.doi.org/10.1186/s40536-016-0025-3}.

\bibitem[{Vania et~al.(2021)Vania, Htut, Huang, Mungra, Pang, Phang, Liu, Cho,
  \protect\BIBand{} Bowman}]{vania2021compare}
Vania C, Htut PM, Huang W, Mungra D, Pang RY, Phang J, Liu H, Cho K, Bowman SR
  (2021) Comparing test sets with item response theory. \emph{Proceedings of
  the Association for Computational Linguistics} (Association for Computational
  Linguistics).

\bibitem[{Wu et~al.(2020)Wu, Davis, Domingue, Piech, \protect\BIBand{}
  Goodman}]{wu2020virt}
Wu M, Davis R, Domingue B, Piech C, Goodman ND (2020) Variational item response
  theory: Fast, accurate, and expressive. \emph{13th International Conference
  on Educational Data Mining}.

\bibitem[{Zhang et~al.(2021)Zhang, Chen, Gendreau, \protect\BIBand{}
  Langevin}]{zhang2021LearningBased}
Zhang X, Chen L, Gendreau M, Langevin A (2021) Learning-{{Based
  Branch-and-Price Algorithms}} for the {{Vehicle Routing Problem}} with {{Time
  Windows}} and {{Two-Dimensional Loading Constraints}}. \emph{INFORMS Journal
  on Computing} ISSN 1091-9856,
  \urlprefix\url{http://dx.doi.org/10.1287/ijoc.2021.1110}.

\end{thebibliography}

\end{document}